\documentclass[letterpaper]{article} 
\usepackage{aaai24}  
\usepackage{times}  
\usepackage{helvet}  
\usepackage{courier}  
\usepackage[hyphens]{url}  
\usepackage{graphicx} 
\usepackage{natbib}  
\usepackage{caption} 
\usepackage{algorithm}
\usepackage{algorithmic}
\usepackage{amsmath}
\usepackage{amssymb}
\usepackage{xcolor}
\usepackage{multirow}
\usepackage{booktabs}
\usepackage{xspace}
\usepackage{subcaption}
\newcommand{\alg}{\textbf{\texttt{ProLAD}}\xspace}
\newcommand{\simalg}{\textbf{\texttt{ProLAD-sim}}\xspace}
\newcommand{\lalg}{\textbf{\texttt{ProLAD-loss}}\xspace}
\newcommand{\nalg}{\textbf{\texttt{TAN}}\xspace}
\newcommand{\wnalg}{\textbf{\texttt{TA}}\xspace}

\usepackage{newfloat}
\usepackage{listings}
\DeclareCaptionStyle{ruled}{labelfont=normalfont,labelsep=colon,strut=off} 
\floatstyle{ruled}
\newfloat{listing}{tb}{lst}{}
\floatname{listing}{Listing}
\title{Leveraging Normalization Layer in Adapters With Progressive Learning and \\
Adaptive Distillation for Cross-Domain Few-Shot Learning}

\author {
    Yongjin Yang,
    Taehyeon Kim,
    Se-Young Yun
}
\affiliations {
    KAIST AI \\ 
    \texttt{\{dyyjkd, potter32, yunseyoung\}@kaist.ac.kr}
}

\begin{document}

\maketitle

\begin{abstract}

Cross-domain few-shot learning presents a formidable challenge, as models must be trained on base classes and then tested on novel classes from various domains with only a few samples at hand. 
While prior approaches have primarily focused on parameter-efficient methods of using adapters, they often overlook two critical issues: shifts in batch statistics and noisy sample statistics arising from domain discrepancy variations. 
In this paper, we introduce a novel generic framework that leverages normalization layer in adapters with \underline{Pro}gressive \underline{L}earning and \underline{A}daptive \underline{D}istillation (\alg), marking two principal contributions. 
First, our methodology utilizes two separate adapters: one devoid of a normalization layer, which is more effective for similar domains, and another embedded with a normalization layer, designed to leverage the batch statistics of the target domain, thus proving effective for dissimilar domains.
Second, to address the pitfalls of noisy statistics, we deploy two strategies: a progressive training of the two adapters and an adaptive distillation technique derived from features determined by the model solely with the adapter devoid of a normalization layer. Through this adaptive distillation, our approach functions as a modulator, controlling the primary adapter for adaptation, based on each domain.
Evaluations on standard cross-domain few-shot learning benchmarks confirm that our technique outperforms existing state-of-the-art methodologies.

\end{abstract}
\section{Introduction}

Deep neural networks, while showing remarkable aptitude for visual recognition tasks when trained on large datasets, face a significant challenge when applied to real-world scenarios characterized by diverse domains. 
The challenges become even more pronounced when these networks encounter unseen data with scarce samples during operational deployment. 
This is where Cross-Domain Few-Shot Learning (CD-FSL) comes into play.
CD-FSL aims to leverage knowledge from a source domain (e.g., ImageNet \cite{deng2009imagenet}), where abundant labeled data is available, to learn a predictive model for a target domain (e.g., Fungi \cite{schroeder2018fgvcx}) where labeled data is scarce. The core objective of this learning paradigm is to cultivate a model that can generalize effectively to new classes within the target domain using a limited set of examples. 

\begin{figure}[ht]
    \centering
  \includegraphics[width=1.0\columnwidth]{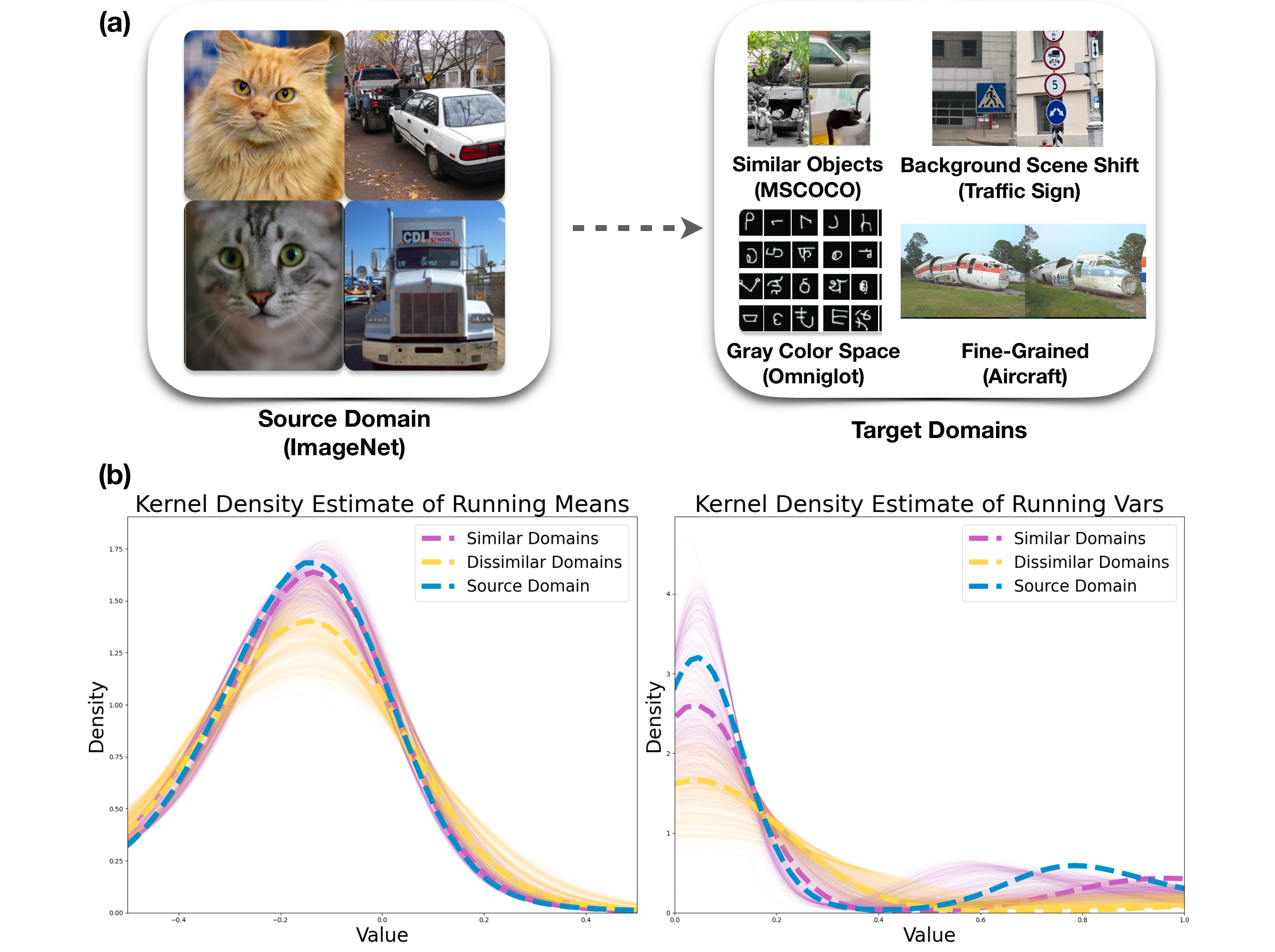}
  \caption{
  (a) Domain shift encountered during deployment in CD-FSL, (b) Kernel density estimation of batch statistics across different domains, where bold dotted lines represent the mean value of batch statistics for each domain and light solid lines represent the batch statistics of individual samples, showing high variations. Domain similarity is computed using Earth-Mover Distance (EMD) as outlined in \cite{cui2018large, oh2022understanding}, and the top seven similar domains are identified as similar domains. Detailed information is described in Appendix~\ref{appendix:domainsim}.
  }
    \label{fig:1}
\end{figure}

The primary challenges encountered in CD-FSL stem from the diverse domain discrepancies \cite{li2020rethinking, oh2022understanding} and the scarcity of samples. 
As illustrated in Figure \ref{fig:1} (a), the model is required to adjust to new domains that may differ in aspects such as color space and background scenes, focal features essential for accurate classification, while using only a restricted number of samples. 
Adaptation from the source domain to the target domain is challenging since it easily poses a high risk of overfitting to a few samples.
To address this challenge, a parameter-efficient approach that fine-tunes only linear adapters has gained attention in CD-FSL \cite{requeima2019fast, bateni2020improved, li2022cross}. 
This approach aims to balance the need for domain adaptation with the risk of overfitting by limiting the number of parameters to be fine-tuned.

However, relying solely on linear adapters fails to address the distribution shift in latent features, which is induced by statistics in normalization layers biased towards the source set. 
Figure \ref{fig:1} (b) illustrates the batch statistics of the source (ImageNet) and target domains based on domain similarity in the Meta-Dataset \cite{triantafillou2019meta}. 
It is observed that underlying batch statistics in the source and new domains can significantly differ in the case of dissimilar domains that do not share any similar classes with the source set.  
This bias in statistics towards the source set can render the additional learned parameters insufficient for domain adaptation \cite{li2016revisiting, bilen2017universal, du2020metanorm, mirza2022norm}.
Therefore, the consideration of normalization layers in adapter-based approaches is necessary.

CD-FSL faces an additional challenge where attempting to incorporate statistics from target domain with a few samples may result in noisy batch statistics. 
This is evident in Figure \ref{fig:1} (b), as highlighted by the light solid lines, complicating the accurate representation of a domain statistics.
This challenge becomes more pronounced for target domains similar to the source, where the statistics of the target domains substantially overlap with those of the source. As a result, integrating noisy data can disrupt the well-established statistics from the pre-trained batch normalization layers.
Conversely, for dissimilar domains, even slightly noisy statistics better represent the statistics of target domain than the statistics of source. 
Given this scenario, there remains a demand for a method that can dynamically leverage statistics based on domain similarity considering the nature of diverse test domains of CD-FSL.

This paper presents a novel generic framework which leverages normalization layers in adapters with \underline{Pro}gressive \underline{L}earning and \underline{A}daptive \underline{D}istillation (\alg) that progressively trains adapters in two stages with adaptive distillation.
Two types of adapters are utilized: Task Adapter with Normalization (\nalg), which employs normalization layer that collects batch statistics, and Task Adapter (\wnalg), which solely comprises a linear adapter. 
The \nalg is optimized for dissimilar domains due to its normalization layer, while the \wnalg is best suited for similar domains, ensuring that well-represented batch statistics remain undisturbed. 
During the adapter training process, our method implement a progressive learning approach, training \wnalg followed by \nalg sequentially, and integrate adaptive distillation. 
This adaptive distillation, informed by the estimation of domain similarity, controls the activation of each adapter. 
Consequently, the model leans on \wnalg for similar domains and primarily engages \nalg when faced with dissimilar domains.
We demonstrate that our \alg outperforms baseline methods on the widely used benchmark Meta-Dataset, especially in more challenging settings where the feature extractor is trained on a single domain dataset (ImageNet) and then tested on all the other domains.

\section{Related Work}
\label{sec:related_work}

\subsection{Cross-Domain Few-Shot Learning}

The early algorithms for solving the cross-domain few-shot classification problem are based on meta-learning, which can be categorized into three categories: model-based \cite{santoro2016meta, munkhdalai2017meta, zhmoginov2022hypertransformer}, optimization-based \cite{finn2017model, antoniou2018train}, and metric-based \cite{vinyals2016matching, snell2017prototypical}. However, recent studies have demonstrated that fine-tuning methods outperform meta-learning methods on few-shot classification tasks \cite{chen2019closer, dhillon2019baseline, chen2021meta, tian2020rethinking, chowdhury2021few}, especially for cross-domain few-shot learning \cite{guo2020broader}, since meta-learning is more prone to overfitting on base classes \cite{dumoulin2021comparing}.

The common approach for cross-domain few-shot learning is to learn task-agnostic feature extractors using the base classes and then fine-tune with task-specific weights to handle the novel classes \cite{requeima2019fast, dvornik2020selecting, tao2022powering, bateni2020improved, li2021universal, li2022cross, triantafillou2021learning, liu2021multi}. These task-specific weights are decided using an auxiliary system \cite{dvornik2020selecting, liu2020universal, triantafillou2021learning, requeima2019fast, bateni2020improved, liu2021multi}, or through training from scratch regardless of the task \cite{tao2022powering, li2021universal, li2022cross}. Recent methods have utilized adapters as task-specific weights to efficiently balance between domain adaptation and overfitting \cite{requeima2019fast, bateni2020improved, li2022cross}. These methods either adopt FiLM layers \cite{perez2018film} as adapters \cite{requeima2019fast, bateni2020improved} or residual adapters parameterized by matrix \cite{li2022cross}.


\begin{figure*}[htp]
    \centering
    \includegraphics[width=1.0\textwidth]{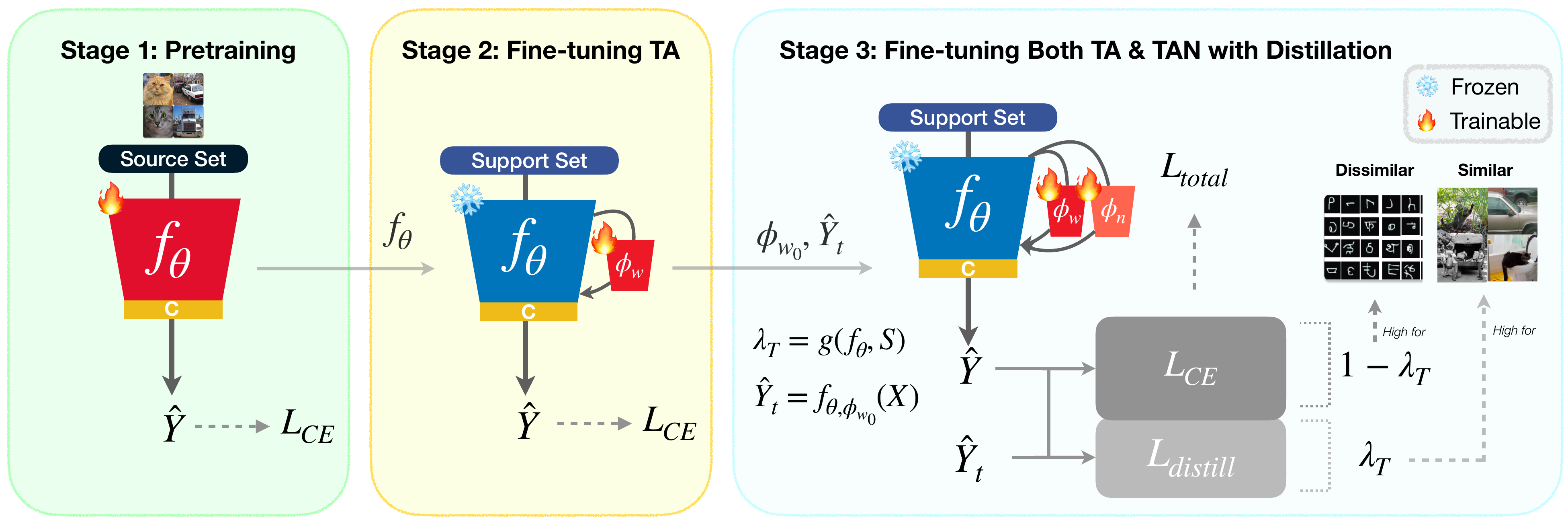}
    \caption{Overview of our method. In stage 1, the feature extractor is pretrained on the source dataset. In stage 2, an adapter \wnalg is fine-tuned using the support set data, while the backbone remains frozen. In stage 3, with \wnalg initialized from the outcomes of stage 2 and adaptive distillation applied using the logits from the stage 2 model for teacher prediction, both adapters are further fine-tuned on the support set. The classifier $c$ is reinitialized at the beginning of each stage.  The adaptive coefficient, $\lambda_{\mathcal{T}}$, varies based on domain similarities.}
    \label{fig:2}
\end{figure*}

\subsection{Knowledge Distillation for Few-shot Learning}

Knowledge distillation is a popular technique in deep learning that transfers the knowledge from a large, pre-trained model (teacher) to a smaller, more efficient model (student) \cite{hinton2015distilling}. Knowledge distillation has also been applied to few-shot learning \cite{tian2020rethinking, rajasegaran2020self}, primarily using self-distillation with identical architectures. Unlike previous approaches, we demonstrate that the knowledge distillation can be useful even during the fine-tuning stage with similar architectures, serving as a regularizer for few-shot training.

\section{Method}
\label{sec:method}

This section presents a novel framework, termed Leveraging Normalization Layer in Adapters with Progressive Learning and Adaptive Distillation (\alg), designed for CD-FSL. Our approach incrementally trains two adapters, \wnalg and \nalg with adaptive distillation. An overview of our method is depicted in Figure \ref{fig:2}, and the detailed pseudo-algorithm is provided in Appendix~\ref{appendix:algorithm}.

\subsection{Training Adapters with Progressive Learning and Distillation}

\subsubsection{Design of Adapters}
\label{method:design_adapters}

\begin{figure}[t]
    \centering
    \includegraphics[width=1.0\columnwidth]{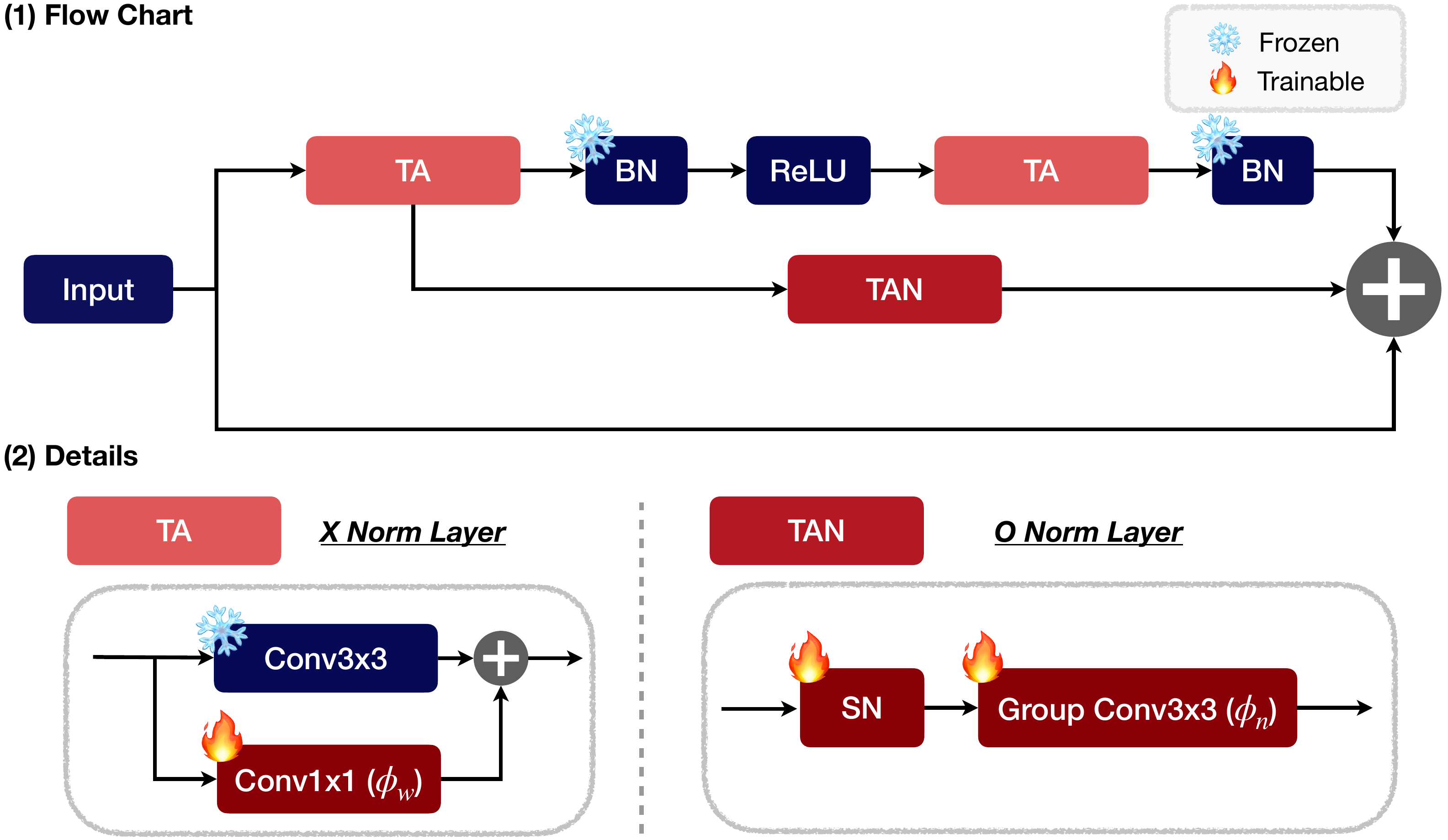}
    \caption{The entire structure of our adapters for each ResNet block. \wnalg\space is $1\times1$ convolutional adapter parameterized by matrix. \nalg\space consists of normalization layer (SN) without affine layer followed by group convolutional layer.}
    \label{fig:3}
\end{figure}

Before delving into each stage, we first provide an overview of the two adapters, \nalg and \wnalg, as depicted in Figure \ref{fig:3}.
First adapter, \wnalg, parameterized by $\phi_w$, employs $1\times1$ convolutional adapter embedded within each $3\times3$ convolutional layer without a normalization layer, following \citet{li2022cross}. 
The role of \wnalg is pivotal in adapting to similar domains, as it does not interfere with the well-represented batch statistics.

In contrast, \nalg, parameterized by $\phi_n$, is equipped with a Standard Normalization layer (SN), which is a batch normalization layer without affine layer that integrates the statistics of the target domain, combined with a group convolutional layer.
The inclusion of this normalization layer in \nalg is crucial for domain adaptation in dissimilar domains, as it gathers additional statistics from the target domain.
Furthermore, we introduce an extra group convolution layer to accommodate larger receptive fields with fewer parameters, replacing the role of an affine layer. 
Detailed insights into these design choices are explored in Section~\ref{ex:ablation_design_adapters}.

The effects of these adapters are summarized in Table \ref{table:summary_adapters}.
More detailed results are presented in Section \ref{ex:ablation_role_adapters}.
As observed, \nalg excels in dissimilar domains (+4.1\%), highlighting the efficacy of the normalization layer in handling low domain similarity. 
This finding supports that incorporating the batch statistics of the target domain is essential for dissimilar domains. 
Conversely, \wnalg performs better in similar domains (+2.8\%), suggesting that the noisy nature of statistics can hinder the training process for high domain similarity. 
This observation paves the way for the introduction of an adaptive mechanism designed to amplify the activation of \nalg for dissimilar domains and \wnalg for similar ones, which will be discussed subsequently.

\begin{table}[ht]
\centering
{\small
\renewcommand{\arraystretch}{1.1}
\resizebox{0.60\columnwidth}{!}{
\begin{tabular}{lcc}
\toprule[1pt]
{Adapter} & {\begin{tabular}[c]{@{}c@{}}Avg.\\[-0.5ex]
Similar \end{tabular}} & {\begin{tabular}[c]{@{}c@{}}Avg.\\[-0.5ex]
Dissimilar \end{tabular}} \\
\midrule
\wnalg & 68.0 & 81.7\\
\nalg & 65.2& 85.8 \\
\bottomrule[1pt]
\end{tabular}
}}
\caption{Summary of accuracy using each proposed adapter based on domain similarity.}
\label{table:summary_adapters}
\end{table}

\subsubsection{Step-by-Step Methods}
\label{method:step-by-step}
As discussed, training one of the two adapter choices is advantageous either for similar or dissimilar domains, but not both. This necessitates an adaptive method to function as a switch, allowing us to selectively employ primary adapters based on domain similarity. Motivated from this insight, our method comprises three distinct training stages with these adapters.

In stage 1, we pretrain the feature extractor, parameterized by $\theta$, using source data, as a transfer learning approach has been demonstrated to be effective for CD-FSL \cite{guo2020broader}. Formally, let $D_b = \{(\mathbf{x}, y)\}$  be a source dataset comprising image and label pairs. We train the model on this source dataset using the cross-entropy loss, defined with dataset $D$ and embedding function $f$, as follows:

\begin{equation}
\label{eq:ce_loss}
L_{CE}(f,D) = -\sum_{\mathbf{x}, y \in D } y \log(f(\mathbf{x}))
\end{equation}

\noindent For pretraining, we use $L_{CE}(f_\theta, D_b)$ as a loss function.

In stage 2, our primary objective is to train \wnalg while keeping the feature extractor frozen and then save the features to be used as the teacher, denoted by $\mathbf{\hat{Y}_t}$.
This phase plays a crucial role in steering the subsequent training of \nalg, as both the teacher features and the initialized parameters of \wnalg are determined at this stage.
In particular, let $S = \{(\mathbf{x_i}, y_i)\}_{i=1}^{|S|}$ be a support set consisting of $|S|$ image-label pairs, which is divided into $\mathbf{X}=\{\mathbf{x_i}\}_{i=1}^{|S|}$ and $Y=\{y_i\}_{i=1}^{|S|}$. We train \wnalg using $L_{CE}(f_{\theta, \phi_w}, S)$ as a loss function. Then, $\mathbf{\hat{Y}_t}$ is calculated with $\mathbf{\hat{Y}_t}=f_{\theta, \phi_w}(X)$.

In stage 3, we proceed to train both \nalg and \wnalg, initializing \wnalg with the values trained in the second stage, denoted as $\phi_{w_0}$, to set a starting point for further training.
Moreover, we apply distillation using the teacher $\mathbf{\hat{Y}_t}$. 
This distillation is modulated based on domain similarities by an adaptive coefficient $\lambda_{\mathcal{T}}$. 
This coefficient is determined by the function $g(f_\theta, S)$, giving us an approximation of the domain shift magnitude. 
With this adaptive distillation, our method can select the required adapters depending on domain similarities.
In the case of similar domains, our method prompts a more distillation and hence relies on \wnalg for adaptation since the activation of \nalg is reduced to follow the teacher $\mathbf{\hat{Y}_t}$. 
On the contrary, for dissimilar domains, our method prioritizes the influence of the ground truth one-hot labels during training, applying minimal distillation and consequently allowing \nalg to play a more dominant role in adaptation.
As the classification depends on the cosine distance, distillation loss is computed using cosine similarity defined as:

\begin{equation} \label{eq:cos_sim}
\small
cos(f_{\theta, \phi_n, \phi_w}, X, \hat{Y}_t) = \!\!\!\!\!\!\! \sum_{\mathbf{x} \in X, \hat{y}_t \in  \hat{Y}_t} \!\frac{f_{\theta, \phi_n, \phi_w}(\mathbf{x})^T \mathbf{\hat{y}_t}}{||f_{\theta, \phi_n, \phi_w}(\mathbf{x})||_2 ||\mathbf{\hat{y}_t}||_2}
\end{equation}

\noindent where $L_{distill} = 1 - cos$. As a result, the final training loss for this stage is defined as:

\begin{equation*}
\label{eq:total_loss}
\begin{split}
L_{total} &= (1-\lambda_{\mathcal{T}})\cdot L_{CE}(f_{\theta, \phi_w, \phi_n}, S)  \\ & + \lambda_{\mathcal{T}} \cdot L_{distill}(f_{\theta, \phi_n, \phi_w}, X, \hat{Y}_t)
\end{split}
\end{equation*}

\noindent Subsequently, we explore potential methods to compute $\lambda_{\mathcal{T}}$.

\subsection{Domain Adaptive Coefficient for Distillation}

An adaptive coefficient $\lambda_{\mathcal{T}}$ is essential for modulating the activation of \nalg, thereby selectively utilizing additional target statistics based on domain similarity.
Several approaches could be employed to design the $g$ function to compute $\lambda_{\mathcal{T}}$, aiming to estimate domain similarity without additional parameters.
In the following sections, we will explore two such variants: one based on similarities between features, and the other on performance metrics.
An in-depth exploration of the impact of $\lambda_{\mathcal{T}}$ is detailed in Section \ref{ex:ablation_adaptive_coefficient}.

\subsubsection{Difference between Inter-class and Feature Similarities}

To estimate domain similarity, we utilize embedded feature similarities. Two metrics are defined in classification tasks: intra-class similarity, which measures the similarity of features within each class, and inter-class similarity, which evaluates the similarity between class prototypes. The distinction between these similarities helps determine how individual class features cluster relative to the entire feature distribution, as a larger difference in these similarities implies a more pronounced domain similarity. However, for classes represented by a single sample, computing the intra-class similarity is not feasible. There, our method uses feature similarities, which measure the similarity across all feature pairs, as an alternative. Hence, our method estimates domain similarity by comparing the distinctness of class prototypes to the similarity between pairs of individual samples.

To determine both types of similarities, we compute the cosine similarities separately. Let $cos^{feat}_{ij}$ denote the cosine similarity between the i-th and j-th features, and $cos^{inter-cls}_{ij}$ denote the similarity between the class prototypes of the i-th and j-th classes. Each cosine similarity is evaluated using Eq.~\eqref{eq:cos_sim}. The adaptive distillation coefficient is derived by considering the difference between the computed similarities, as follows:

\begin{equation} \label{eq:coeff_sim}
\small
\lambda_{\mathcal{T}} \!=\! exp(\beta\cdot(|\frac{1}{N_s}\cdot\sum_{i<j}cos^{feat}_{ij}\! -\! \frac{1}{N_c}\cdot\sum_{i<j}cos^{inter-cls}_{ij}|  - \!1 ))
\end{equation}

\begin{figure}[t]
    \centering
    \includegraphics[width=1.0\columnwidth]{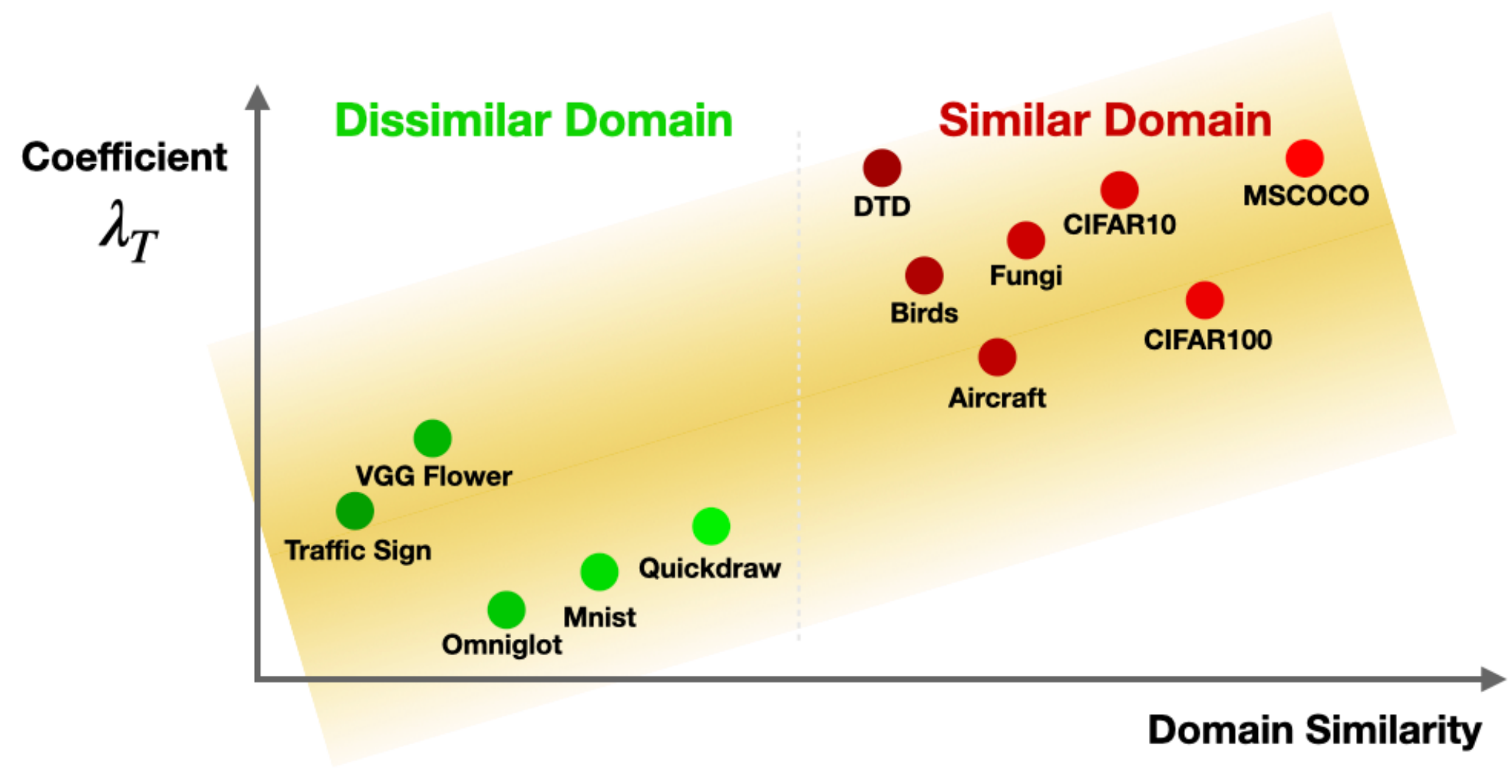}
    \caption{Correlation between our adaptive coefficient using similarity difference and domain similarity using EMD.}
    \label{fig:4}
\end{figure}

\noindent In this equation, $\beta$ is a scaling factor, $N_s$ denotes the number of feature pairs from support set and $N_c$ represents the number of class pairs.  We employ the exponential function due to its value range being constrained between zero and one. According to this equation, a larger difference between the two similarity measures results in a higher $\lambda_{\mathcal{T}}$, which is favorable for similar domains as intended. Coefficients calculated with Eq.~\eqref{eq:coeff_sim} exhibit a high correlation with domain similarity, as illustrated in Figure \ref{fig:4}, showcasing the efficacy of this metric as a reliable estimator of domain similarity. We refer to this technique as \simalg.

\subsubsection{Support Set Loss \& Accuracy}

Another metric to estimate domain similarity utilizes performance indicators from the pretrained feature extractor on the support set, such as loss and accuracy. These metrics are especially relevant in understanding task distribution shift \cite{luo2022channel, luo2023closer} or task difficulty \cite{oh2022understanding}, both correlating with domain shift. A closer domain similarity is indicated by a lower loss and higher accuracy. Specifically, let $acc$ denote the accuracy and $L_{orig} = L_{CE}(f_\theta, S)$ be the loss on the support set $S$, where the loss is computed using Eq.~\eqref{eq:ce_loss}. The adaptive coefficient is then given by:

\begin{equation} \label{eq:coeff_loss}
\lambda_{\mathcal{T}} = exp\left(-\beta \cdot L_{orig} \cdot (1-acc)\right)
\end{equation}

\noindent As for \simalg, we employ the exponential function with $\beta$ as a scaling factor. This equation implies that higher losses and lower accuracies yield lower coefficients suited for dissimilar domains, while the inverse leads to higher coefficients optimal for similar domains. We term this approach \lalg.

\section{Experiments}
\label{sec:experiments}
\subsection{Experimental Setup}
\paragraph{Datasets} We utilize the Meta-Dataset \cite{triantafillou2019meta}, which is a CD-FSL benchmark consisting of 13 different datasets. We evaluate our method using the standard setting as described in the original paper. For the Single-Domain Learning (SDL) setting, we first train the feature extractor on ImageNet and then test it on all 13 datasets in the Meta-Dataset. On the other hand, for the Multi-Domain Learning (MDL) setting, we train the feature extractor on 8 datasets and then evaluate its performance on all 13 datasets. Mean accuracy over the 600 episodes are reported with 95 confidence interval. For the analysis of domain similarities in the SDL setting, we assess domain similarities using the EMD distance \cite{rubner1998metric}. We classify Traffic Sign, VGG Flower, MNIST, Omniglot, and Quick Draw as the five most dissimilar domains, while the others are deemed similar. Details are in Appendix~\ref{appendix:domainsim}.

\paragraph{Implementation Details} We use the standard ResNet18 \cite{he2016deep} model as the feature extractor, which is a commonly used architecture for CD-FSL \cite{bateni2020improved, dvornik2020selecting, liu2020universal, li2021universal}. For the SDL setting, we adopt the training procedure proposed in \citet{dvornik2020selecting, li2022cross}. In the MDL setting, we use Universal Representation Learning~(URL) \cite{li2021universal} to train on multiple domains to train across multiple domains, thereby avoiding negative transfer.

For fine-tuning adapters, we utilize the Adadelta optimizer \cite{zeiler2012adadelta} with a learning rate of 0.5. We set the scaling coefficient $\beta$ in distillation to the value of 1.5 for both variants of \alg. We use a large momentum size of 0.8 for normalization layer because the feature values can change significantly between epochs. Additionally, we employ a group size of 8 for the group convolutional network. 
More details can be found in the Appendix~\ref{appendix:imdetails}.

\paragraph{Baselines} 
For the baselines, we compare our methods to state-of-the-art methods, including standard fine-tuning and Protonet-based methods \cite{triantafillou2019meta}. We also consider adapters with meta-net such as Simple CNAPS \cite{bateni2020improved}, URT \cite{liu2020universal}, FLUTE \cite{triantafillou2021learning}, tri-M \cite{liu2021multi}, Transductive CNAPS \cite{bateni2022enhancing}, selective method like SUR \cite{dvornik2020selecting}, hyperparameter-optimization-based method of BOHB \cite{saikia2020optimized}, and simple linear adapter-based methods like TSA \cite{li2022cross} and URL \cite{li2021universal}. For TSA \cite{li2022cross} in the SDL setting, we have reproduced the results ourselves, incorporating a data shuffling modification.

\subsection{Main Results}
\label{ex:main_results}

\subsubsection{Single-Domain Learning (SDL)}
\label{ex:sdl}
\begin{table*}[t]
\addtolength{\tabcolsep}{-2.0pt}\small
\centering
\small
\begin{tabular}{cccccccc|cc}
\toprule
        ~ & Finetune  & Protonet  & {\begin{tabular}[c]{@{}c@{}}Proto\\[-0.5ex]
-MAML \end{tabular}} & BOHB  & {\begin{tabular}[c]{@{}c@{}}Simple\\[-0.5ex]
CNAPS  \end{tabular}}& {\begin{tabular}[c]{@{}c@{}}FLUTE\\[-0.5ex] \end{tabular}}  & TSA  & \textbf{\lalg} & \textbf{\simalg} \\
        \midrule
        ImageNet & 45.8 ± 1.1 &  50.5 ± 1.1 & 49.5 ± 1.1 & 51.9 ± 1.1 & 54.8 ± 1.2 & 46.9 ± 1.1 & 57.1 ± 1.2 & \textbf{57.8 ± 1.2}  &  {57.2 ± 1.1}\\
        \midrule
        Omniglot & 60.9 ± 1.6 & 60.0 ± 1.4 & 63.4 ± 1.3 & 67.6 ± 1.2 & 62.0 ± 1.3 & 61.6 ± 1.4 & 76.2 ± 1.2 &  82.7 ± 1.1   & \textbf{84.1 ± 1.2}\\ 
        Aircraft & 68.7 ± 1.3 & 53.1 ± 1.0 & 56.0 ± 1.0 & 54.1 ± 0.9 & 49.2 ± 0.9 & 48.5 ± 1.0 & 71.9 ± 1.0 & 74.8 ± 1.2  & \textbf{76.1 ± 1.2}\\ 
        Birds & 57.3 ± 1.3 & 68.8 ± 1.0 & 68.7 ± 1.0 & 70.7 ± 0.9 & 66.5 ± 1.0 & 47.9 ± 1.0 & 74.4 ± 0.9 & 74.1 ± 1.0  & \textbf{75.5 ± 1.0} \\ 
        Textures & 69.0 ± 0.9 & 66.6 ± 0.8 & 66.5 ± 0.8 & 68.3 ± 0.8 & 71.6 ± 0.7 & 63.8 ± 0.8 & 76.9 ± 0.7 & 77.1 ± 0.8  & \textbf{77.7 ± 0.8} \\ 
        Quick Draw & 42.6 ± 1.2 & 49.0 ± 1.1 & 51.5 ± 1.0 & 50.3 ± 1.0 & 56.6 ± 1.0 & 57.5 ± 1.0 & 66.4 ± 0.9 & 69.6 ± 1.0  & \textbf{70.6 ± 1.0} \\ 
        Fungi & 38.2 ± 1.0 & 39.7 ± 1.1 & 40.0 ± 1.1 & 41.4 ± 1.1 & 37.5 ± 1.2 & 31.8 ± 1.0 & 46.7 ± 1.1 & 46.1 ± 1.1   & \textbf{46.8 ± 1.2} \\ 
        VGG Flower & 85.5 ± 0.7 & 85.3 ± 0.8 & 87.2 ± 0.7 & 87.3 ± 0.6 & 82.1 ± 0.9 & 80.1 ± 0.9 & 91.2 ± 0.5 &  92.2 ± 0.6   &  \textbf{92.9 ± 0.6}\\ 
        Traffic Sign & 66.8 ± 1.3 & 47.1 ± 1.1 & 48.8 ± 1.1 & 51.8 ± 1.0 & 63.1 ± 1.1 & 46.5 ± 1.1 & 81.9 ± 1.0 & \textbf{89.6 ± 0.9} & 89.4  ± 0.9 \\ 
        MSCOCO & 34.9 ± 1.0 & 41.0 ± 1.1 & 43.7 ± 1.1 & 48.0 ± 1.0 & 41.4 ± 1.0 & 45.8 ± 1.0 & \textbf{55.7 ± 1.0} &  54.6 ± 1.1   & 55.4 ± 1.1 \\ 
        MNIST & - & - & - & - & - & 80.8 ± 0.8 & 92.9 ± 0.6 &  95.5 ± 0.5  &  \textbf{95.8 ± 0.5} \\ 
        CIFAR-10 & - & - & - & - & - & 65.4 ± 0.8 & 79.4 ± 0.7 &  79.2 ± 0.8  & \textbf{79.7 ± 0.8} \\ 
        CIFAR-100 & - & - & - & - & - & 52.7 ± 1.1 & \textbf{70.9 ± 1.0} & 70.1 ± 0.9      & 70.3 ± 1.0 \\ 
        \midrule
        Avg. seen & 45.8 & 50.5 & 49.5 & 51.9 &  54.8  & 46.9 & 57.1 & \textbf{57.8}  & {57.2} \\ 
        Avg. unseen & 58.2 & 56.7 & 58.4 & 60.0 & 59.3  & 53.2 & 73.7 & {75.4}  & \textbf{76.2} \\
        \midrule
        Avg. similar & 53.6 & 53.8 & 55.0 & 56.5 & 53.2  & 51.0 & 68.0 & {68.0}  & \textbf{68.8} \\
        Avg. dissimilar & 64.0 & 60.4 & 62.7 & 64.3 & 66.0  & 65.3 & 81.7 & {85.9}  & \textbf{86.6} \\
        \midrule
        Avg. all & 57.0 & 56.1 & 57.5 & 59.2 &  58.9  & 52.6 & 72.4 & 74.0  & \textbf{74.7} \\
        \bottomrule
    \end{tabular}
\caption{Comparison to state-of-the-art methods on Meta-Dataset with the single-domain setting where the feature extractor is trained only on ImageNet and then test on all datasets. Mean accuracy, 95 confidence interval are reported. } 
\label{table:sdl}
\end{table*}

Table \ref{table:sdl} presents the results of our proposed method on the Meta-Dataset with the SDL setting. A feature extractor is trained on the ImageNet dataset and evaluation is carried out on all datasets, including ImageNet, with different classes. Both of our methods surpass other techniques in terms of average accuracy. Specifically, for the majority of domains, our methods excel: \lalg performs better in 8 out of 13 domains, and \simalg in 11 out of 13. Significant improvements are observed in domains distinctly dissimilar to ImageNet (+4.9\%), such as Omniglot (+7.9\%) and Traffic Sign (+7.7\%), attributed to the high activation of \nalg. For domains closely related to the base domain, like CIFAR-10 (+0.5\%), CIFAR-100 (-0.6\%), and MSCOCO (-0.3\%), our method demonstrates performance comparable to TSA \cite{li2022cross} (+0.8\%), which trains \wnalg from scratch, showing that our method primarily utilizes \wnalg for similar domains.

Additionally, for most domains, \simalg that employs Eq.~\eqref{eq:coeff_sim} for coefficient demonstrates superior performance over \lalg, which utilizes Eq.~\eqref{eq:coeff_loss} for its coefficient. This performance indicates a more favorable scenario for computing $\lambda_{\mathcal{T}}$: a stronger correlation with domain similarity over other considerations. While loss and accuracy on the support set correlate with domain similarity, they more closely align with the notion of task-level shift \cite{luo2022channel, luo2023closer} and task difficulty \cite{oh2022understanding}. However, variations in similarities directly correlate with domain similarity, as illustrated in Figure \ref{fig:4}. Given the evident superiority of \simalg over \lalg, our subsequent experiments will primarily focus on \simalg.

\subsubsection{Multi-Domain Learning (MDL)}
\label{ex:mdl}
\begin{table*}[ht]
\centering
\small
\begin{tabular}{ccccccccc|cc}
    \toprule
        ~ & {\begin{tabular}[c]{@{}c@{}}Simple\\[-0.5ex]
CNAPS  \end{tabular}} &  {\begin{tabular}[c]{@{}c@{}}Transductive\\[-0.5ex]
CNAPS  \end{tabular}}  & SUR & URT  & FLUTE  & tri-M  & URL & TSA  & \lalg & \simalg \\ 
        \midrule
        Avg. seen &  74.6 & 75.1 & 75.2 & 76.7 & 76.2 & 74.5 & 80.0& 80.2 & 80.6 &\textbf{80.9}  \\ 
        Avg. unseen & 65.8 & 66.5 & 63.1 & 62.2 & 69.9 & 72.9 & 69.3 & 77.2 & 78.2&\textbf{78.6}  \\ 
        Avg. all &  71.2 & 71.8 & 70.5 & 71.1 & 73.8 & 73.9 & 75.9 & 79.0 & 79.7 &\textbf{80.0} \\ 
        \bottomrule
    \end{tabular}
\caption{Comparison to state-of-the-art methods on Meta-Dataset with the multi-domain setting where the feature extractor is trained on 8 datasets and then test on all datasets. Mean accuracy is reported.} 
\label{table:mdl}

\end{table*}

Table \ref{table:mdl} presents the overall results of our proposed method on the Meta-Dataset under the MDL setting. More detailed results for each dataset, along with their respective confidence intervals, is presented in Appendix~\ref{appendix:further_results}. 
The MDL setting is considered less challenging than the SDL setting because the feature extractor is pre-trained on multiple domains, specifically eight domains, which increases the likelihood that some of the domain properties of the test set have already been learned. Nevertheless, both \lalg and \simalg still outperform other methods in both seen and unseen domains, although not to the same degree as in the SDL setting, given the limited presence of domains significantly distinctive from the source. Overall, the results demonstrate the effectiveness of our method in the easier MDL setting as well.

\subsection{Ablation Study}
\label{ex:ablation}

\begin{table*}[h]
\centering
\small
\setlength{\tabcolsep}{2.5pt}
\renewcommand{\arraystretch}{1.1}
\begin{tabular}{lccccccccccccc|ccc}
    \toprule
       {Adapter}  & {\begin{tabular}[c]{@{}c@{}}Image\\[-0.5ex]
-net \end{tabular}} & {\begin{tabular}[c]{@{}c@{}}Omni\\[-0.5ex]
-glot \end{tabular}}& {\begin{tabular}[c]{@{}c@{}}Airc\\[-0.5ex]
-craft \end{tabular}}& {Birds}& {\begin{tabular}[c]{@{}c@{}}Tex\\[-0.5ex]
-tures \end{tabular}}& {\begin{tabular}[c]{@{}c@{}}Quick\\[-0.5ex]
Draw \end{tabular}}& {\begin{tabular}[c]{@{}c@{}}Fun-\\[-0.5ex]
gi \end{tabular}}& {\begin{tabular}[c]{@{}c@{}}VGG\\[-0.5ex]
Flower \end{tabular}}& {\begin{tabular}[c]{@{}c@{}}Traffic\\[-0.5ex]
Sign \end{tabular}} & {\begin{tabular}[c]{@{}c@{}}MS\\[-0.5ex]
-COCO \end{tabular}} & {\begin{tabular}[c]{@{}c@{}}MN\\[-0.5ex]
-IST \end{tabular}}& {\begin{tabular}[c]{@{}c@{}}CIFAR\\[-0.5ex]
-10 \end{tabular}}& {\begin{tabular}[c]{@{}c@{}}CIFAR\\[-0.5ex]
-100 \end{tabular}} & {\begin{tabular}[c]{@{}c@{}}Avg.\\[-0.5ex]
Similar \end{tabular}}& {\begin{tabular}[c]{@{}c@{}}Avg. Dis\\[-0.5ex]
-similar \end{tabular}}& {\begin{tabular}[c]{@{}c@{}}Avg.\\[-0.5ex]
all \end{tabular}}\\ 
        \midrule
         \wnalg &  57.1& 76.2& 71.9& 74.4& 76.9& 66.4& 46.7 & 91.2& 81.9& \textbf{55.7} & 92.9& 79.4& \textbf{70.9} & 68.0 & 81.7 & 72.4 \\ 
         \nalg &  55.2 & 84.1& 73.2 & 71.6 & 75.7 & 69.0 & 42.5 & 91.2 & \textbf{89.8} & 50.9 & 94.9 & 76.2 & 66.6 & 65.2 & 85.8 & 72.4    \\
         \nalg\space + \wnalg & 55.4 & 83.4 & 72.1& 71.7 & 74.3& 69.0 & 42.4 & 91.6 & 89.7 & 49.6 & 95.7 & 75.1& 65.9& 64.4 & 85.9 & 72.0    \\
         \midrule
         \simalg & \textbf{57.2}& \textbf{84.1}& \textbf{76.1}& \textbf{75.5}& \textbf{77.7} &\textbf{70.6}& \textbf{46.8}& \textbf{92.9}& 89.4& {55.4} & \textbf{95.8}& \textbf{79.7}& 70.3 & \textbf{68.8} &\textbf{86.6}&\textbf{74.7}  \\ 
        \bottomrule
    \end{tabular}
\caption{The ablation study on the adapters in the SDL setting. The term `Adapter' refers to the combination of adapters trained from scratch.} 
\label{table:ablation_adapter_role}
\end{table*}

\begin{table*}[h]
\centering
\small
\setlength{\tabcolsep}{2.5pt}
\renewcommand{\arraystretch}{1.1}
\begin{tabular}{lccccccccccccccc}
    \toprule
       {Method}  & {\begin{tabular}[c]{@{}c@{}}Image\\[-0.5ex]
-net \end{tabular}} & {\begin{tabular}[c]{@{}c@{}}Omni\\[-0.5ex]
-glot \end{tabular}}& {\begin{tabular}[c]{@{}c@{}}Airc\\[-0.5ex]
-craft \end{tabular}}& {Birds}& {\begin{tabular}[c]{@{}c@{}}Tex\\[-0.5ex]
-tures \end{tabular}}& {\begin{tabular}[c]{@{}c@{}}Quick\\[-0.5ex]
Draw \end{tabular}}& {Fungi}& {\begin{tabular}[c]{@{}c@{}}VGG\\[-0.5ex]
Flower \end{tabular}}& {\begin{tabular}[c]{@{}c@{}}Traffic\\[-0.5ex]
Sign \end{tabular}} & {\begin{tabular}[c]{@{}c@{}}MS\\[-0.5ex]
-COCO \end{tabular}} & {MNIST}& CIFAR10& CIFAR100 &{\begin{tabular}[c]{@{}c@{}}Avg.\\[-0.5ex]
all \end{tabular}}\\

        \midrule
         S &  55.4 & 83.4 & 72.1& 71.7 & 74.3& 69.0 & 42.4 & 91.6 & 89.7 & 49.6 & 95.7 & 75.1& 65.9& 72.0\\
         P &   56.2& 83.3& 74.3 & 71.2& 73.8& 69.7& 43.4 & 91.4 & \textbf{90.7} & 50.7 & 95.5 & 73.9& 66.8& 72.4   \\ 
         D + A  & 56.6& 82.7& 74.5 & 74.1& 76.8& 69.5 & 45.5 & 92.3 & 89.7 & 54.7 & \textbf{95.9} & 79.3 & 69.3 & 73.9   \\
         \midrule
         P + D + A (ours) & \textbf{57.2}& \textbf{84.1}& \textbf{76.1}& \textbf{75.5}& \textbf{77.7} &\textbf{70.6}& \textbf{46.8}& \textbf{92.9}& 89.4& \textbf{55.4} & {95.8}& \textbf{79.7}& \textbf{70.3} &\textbf{74.7}  \\ 
        \bottomrule
    \end{tabular}
\caption{Results from the ablation study examining progressive learning and adaptive distillation on the Meta-Dataset within the SDL setting. `S' denotes training both adapters from scratch, `P' represents progressive learning, `D' signifies distillation, while `A' corresponds to adaptive coefficient calculated with Eq.~\eqref{eq:coeff_sim}.} 
\label{table:ablation_training}
\end{table*}

\begin{figure}[t]
    \centering
    \includegraphics[width=1.0\columnwidth]{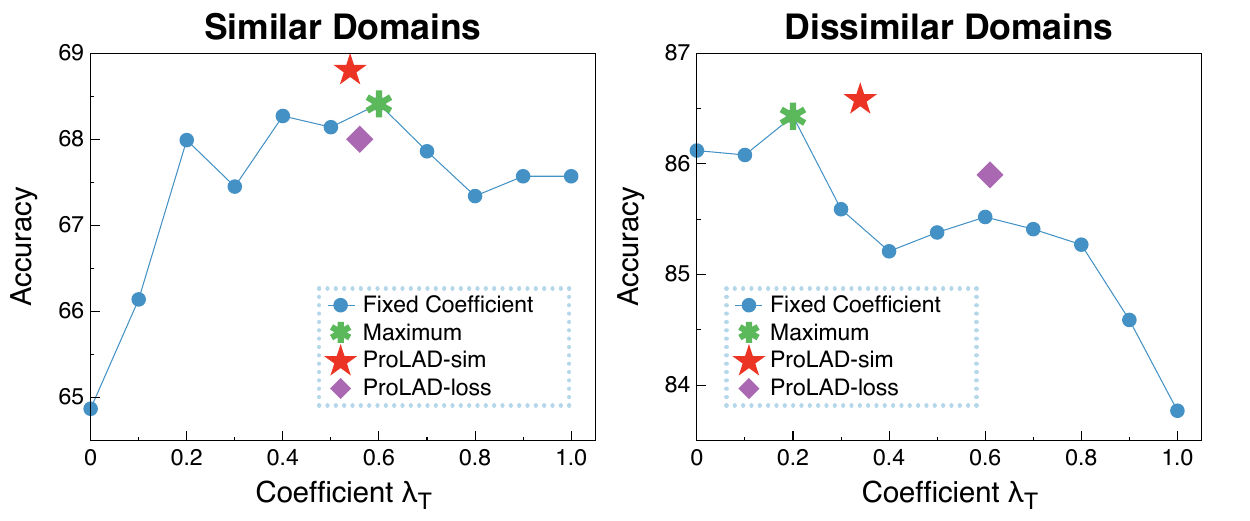}
    \caption{Performance by coefficient $\lambda_{\mathcal{T}}$ based on domain similarities. The green mark denotes the maximum value point with a fixed coefficient. The red and purple marks represent the performance using Eq.~\eqref{eq:coeff_sim} and Eq.~\eqref{eq:coeff_loss}, respectively, where the value of x is the average coefficient across all samples. Our estimated coefficient value closely aligns with the ideal coefficient of a fixed value.}
    \label{fig:5}
\end{figure}

\subsubsection{Performance of Each Adapter}
\label{ex:ablation_role_adapters}

Table \ref{table:ablation_adapter_role} shows the performance of each adapter in the SDL setting. 
\wnalg excels in domains closely related to the base domains (+2.6\%), whereas \nalg performs better on tasks with low domain similarity (+4.1\%). 
These findings corroborate our hypothesis: while incorporating batch statistics from the target domain aids performance in dissimilar domains, it might be less beneficial for similar domains due to the noisy statistics from few-shot scenarios. 
Furthermore, training both adapters from scratch yields performance on par with \nalg, suggesting that the activation strength of \nalg surpasses that of \wnalg. 
Thus, in the absence of distillation, activation of \nalg predominantly serves the adaptation, highlighting that distillation can guide the selection of primary adapters during fine-tuning. 
Our approach exhibits performance akin to \wnalg in similar domains and mirrors \nalg in dissimilar domains, implying that our method effectively increases the activation of optimal adapter based on the domain similarity.

\subsubsection{Progressive Learning and Adaptive Distillation} 
\label{ex:ablation_training}

Table \ref{table:ablation_training} illustrates the impact of progressive learning and adaptive distillation in stage 3. For this analysis, we exclusively employ \simalg, where the coefficient is determined using Eq.~\eqref{eq:coeff_sim}. Collectively, both progressive learning and adaptive distillation emerge as crucial for regularization during the concurrent training of both adapters; this is evidenced by the peak performance exhibited by \alg. Notably, adaptive distillation considerably enhances performance, indicating the pivotal role it plays in dynamically controlling the activation of the two adapters.

\subsubsection{Efficacy of Adaptive Coefficient} 
\label{ex:ablation_adaptive_coefficient} 
Figure \ref{fig:5} illustrates the performance based on a fixed coefficient $\lambda_{\mathcal{T}}$ in conjunction with the average coefficient of \simalg and \lalg. 
For similar domains, the results suggest that while a higher coefficient is generally favorable, an excessively large coefficient leads to decreased performance. 
This decline stems from the potential risks of relying solely on distillation, given the difference architectures of student and teacher.
For dissimilar domains, a lower coefficient is typically preferred as \nalg performs better in these scenarios. 
The coefficient of \simalg closely matches the optimal fixed coefficient and outperforms all fixed coefficients. 
This is because, within the same domain, it can modulate the distillation based on target classes.
In contrast, \lalg calculates $\lambda_{\mathcal{T}}$ similarly to the optimal value for similar domains but differs for dissimilar domains. 
This difference arises because even though these dissimilar domains are distinct from the source, they are relatively easy tasks, resulting in higher accuracy.
Comprehensive results for each dataset can be found in Appendix~\ref{appendix:further_results}.

 \subsubsection{Components of \nalg Adapter} 
 \label{ex:ablation_design_adapters}
Finally, we present an ablation study on the individual components of \nalg when combined with \simalg, as provided in Table \ref{table:adapter_design}. Full results are given in Appendix~\ref{appendix:further_results}. 
Our findings indicate that normalization is pivotal for adaptation; removing SN leads to a decline in performance. Furthermore, our analysis suggests that the FiLM layer is not adequate for feature modulation since its combination with SN does not enhance performance. While the group convolution underperforms on its own, its combination with SN delivers the best performance, underscoring the significance of large receptive fields in feature modification when the features are normalized. In summary, our proposed design surpasses other combinations with a similar number of additional parameters.

\begin{table}[t]
\centering
{\small
\renewcommand{\arraystretch}{1.5}
\resizebox{1.00\columnwidth}{!}{
\begin{tabular}{clc|p{1.2cm}p{1.2cm}|c}
\toprule[1pt]
  \# & {{Architecture}} & {\#params}&{\begin{tabular}[c]{@{}c@{}}Avg.\\[-1ex]
seen \end{tabular}} & {\begin{tabular}[c]{@{}c@{}}Avg.\\[-1ex]
unseen \end{tabular}} & {\begin{tabular}[c]{@{}c@{}}Avg.\\[-1ex]
all \end{tabular}} \\
\hline
(1) & FiLM & {2.38\%} & 57.2 & {74.4}  & {73.0} \\
(2) & Conv & {8.57\%} & 57.2 & {74.8}  & 73.4\\
(3) & GroupConv & {9.35\%} &56.0 & {73.9}  & {72.6}\\
(4) & SN + Conv & {8.57\%} &{56.0} & {75.6}  & {74.1} \\
(5) & SN + FiLM (BN) & {2.38\%} &{57.1} & {74.5}  & {73.1} \\
(6) & SN + GroupConv (Ours) & {9.35\%} & \textbf{57.2} & \textbf{76.2}  & \textbf{74.7} \\
\bottomrule[1pt]
\end{tabular}
}}
\caption{Ablation study on the \nalg components. `Conv' denotes a $1\times1$ convolutional layer, `FiLM' refers to a FiLM layer. `\#params' indicates the number of parameters added for adapters relative to the total parameters in ResNet18.}
\label{table:adapter_design}
\end{table}

\section{Conclusion}

In this paper, we present a method of leveraging normalization layer in adapters with progressive learning and adaptive distillation. Our method employs two distinct adapters: one equipped with a normalization layer that leverages statistics from the target domain, and another without it. These adapters are trained using progressive learning and adaptive distillation. Our approach adeptly addresses the challenges posed by diverse test domains in CD-FSL by selecting the primary adapter based on domain similarity. Experimental results on the standard benchmark, Meta-Dataset, highlight our method's outstanding performance.

\section* {Acknowledgement}
This work was supported by Institute of Information \& communications Technology Planning \& Evaluation (IITP) grant funded by Korea government (MSIT) [No. 2021-0-00907, Development of Adaptive and Lightweight Edge-Collaborative Analysis Technology for Enabling Proactively Immediate Response and Rapid Learning, 90\%] and [No. 2019-0-00075, Artificial Intelligence Graduate School Program (KAIST), 10\%]. We thank Jimin Lee for helping us create figures of our motivation and methods.

\bibliography{aaai24}

\newpage
\appendix

\section{Implementation Details}
\label{appendix:imdetails}

\subsection{Pretraining} 
We adopt the ResNet18 pretraining scheme used in several other studies \cite{dvornik2020selecting, li2022cross}. Specifically, for SDL pretraining, we use an image size of 84x84 and the SGD optimizer with a momentum of 0.9. Data augmentations such as random crops and random color augmentation are applied, along with cosine annealing and a weight decay of $7\times10^{-4}$. For MDL pretraining, we utilize the pretrained URL \cite{li2021universal} with identical hyperparameter settings, maintaining the image size, weight decay, augmentation, and learning rate scheduler as in the SDL setting, with a learning rate of 0.03.

\subsection{Fine-tuning Adapters}
During fine-tuning, \wnalg and \nalg are attached and trained using the Adadelta optimizer \cite{zeiler2012adadelta} with a learning rate of 0.5. Logits are determined using class prototypes based on the cosine distance, bypassing the traditional linear classifier. Instead of the linear classifier, we use the classifier $c$, which represents the preclassifier alignment as described in \citet{li2021universal}. This classifier aligns the features for each class, and is trained with a learning rate of 1.0. For seen domains, both learning rates are reduced by a factor of 0.1. We set the number of iterations to 25 times more after achieving 99\% accuracy on the support set in order to reduce the number of epochs and address the varying number and shots in the Meta-Dataset \cite{triantafillou2019meta}. The momentum is set to 0.8 for standard normalization, and the scale parameter for determining the adaptive distillation coefficient is 1.5 for both variations of \alg. For feature transformation after batch normalization, we utilize 3x3 group convolutional networks with a group size of 8. The parameters of \nalg are initialized with a Gaussian distribution scaled by 1e-4, while those of \wnalg are scaled by 1e-5. For evaluations on the query set, we employ the trained model from stage 3 that incorporates both adapters. All these hyperparameters are same for stage 2 and stage 3

For \simalg, to calculate the similarity between prototypes, we first compute the prototypes using the original feature extractor. We then form a similarity matrix with these prototypes and remove the redundant terms in the upper triangle and diagonal. The average value of the resulting matrix represents our final measure of similarity between the cosine similarities. The similarities between all feature pairs are calculated in a similar manner. For \lalg, we use the pretrained feature extractor without the classifier to compute the cross-entropy loss and accuracy on the support set.

\begin{table*}[h!]
\centering
\small
\setlength{\tabcolsep}{3.5pt}
\renewcommand{\arraystretch}{1.8}
\begin{tabular}{cccccccccccc}
    \toprule
       Omniglot&Aircraft & {Birds}& Textures & {\begin{tabular}[c]{@{}c@{}}Quick\\[-2.5ex]
Draw \end{tabular}}& {Fungi}& {\begin{tabular}[c]{@{}c@{}}VGG\\[-2.5ex]
Flower \end{tabular}}& {\begin{tabular}[c]{@{}c@{}}Traffic\\[-2.5ex]
Sign \end{tabular}} & MSCOCO & {MNIST}& CIFAR-10 & CIFAR-100 \\

        \midrule
         0.846 & 0.860 & 0.856 & 0.851 & 0.849 & 0.862 & 0.835 &  0.835 & 0.871 & 0.838 & 0.863 & 0.875  \\
         
        \bottomrule
    \end{tabular}
\caption{Domain similarities between the domains in Meta-Dataset and the source dataset (ImageNet).}
\label{table:domain_sim}
\end{table*}

\section{Domain Similarity and Batch Statistics}
\label{appendix:domainsim}

\subsection{Domain Similarity}

We utilize the Earth-Mover Distance (EMD) to estimate the similarity between two domains, in the context of transfer learning, as previously explored in \citet{cui2018large, oh2022understanding}. Given two distinct domains $D^A$ and $D^B$ with corresponding class sets $C^A$ and $C^B$, where $C^A \cap C^B = \emptyset$, we can formalize the domain similarity as follows:

\begin{equation} \label{eq:domain_sim}
Sim(D^A, D^B) = \exp(-\gamma EMD(D^A, D^B))
\end{equation}

with  \[EMD(D^A, D^B)) = \frac{\sum_{i\in{C^A}, j\in{C^B}}f_{i,j}d_{i,j}}{\sum_{i\in{C^A}, j\in{C^B}}f_{i,j}}\] 
\noindent where $d_{i,j} = {||p_i - p_j||}_2$, $p_i$ is the class prototype of class $i$, and $f_{i,j}$ is the optical flow obtained by solving the EMD optimization problem. To compute each class prototype, we utilize the pretrained ResNet101 model. While $D^A$ is fixed with ImageNet, which includes classes for training, $D^B$ is obtained from 12 other datasets with classes used for evaluation only. We set $\gamma$ to 0.02 to scale properly for the Meta-Dataset. The resulting domain similarity is presented in Table \ref{table:domain_sim}. In summary, the domain similarities to ImageNet are ranked in the following order: CIFAR-100 $>$ MSCOCO $>$ CIFAR-10 $>$ Fungi $>$ Aircraft $>$ Birds $>$ Textures $>$ Quickdraw $>$ Omniglot $>$ MNIST $>$ VGG Flower $>$ Traffic Sign. We consider domains up to Quickdraw as dissimilar, while we consider CIFAR-100 to TEXTURES as similar domains.

\subsection{Batch Statistics}

In this section, we delve into the methods used for computing batch statistics that are illustrated in Figure \ref{fig:1} (b). For the statistics of the source domain, we extract the values from the first pretrained batch normalization layer of each ResNet block. After combining all the values, we categorize them into bins similar to a histogram and subsequently estimate the distribution using kernel density. For other domains, we sample 100 values and adjust the momentum of batch normalization to 1.0. We then perform a forward pass on the support set and extract the saved running mean and variances, estimating them in a manner consistent with the source statistics. All 1,200 values are presented using solid light lines. For the average value, we compute the mean of statistics from both similar and dissimilar domains, which is then depicted with bold dotted lines.

\section{Algorithm of \alg}
\label{appendix:algorithm}

Algorithm \ref{alg:overview} provides the pseudo-code for our method. For simplicity, we omit the pretraining on the source part and focus solely on the fine-tuning process of stages 2 and 3.

\begin{algorithm}[ht]
\caption{Fine-tuning with Progressive Learning and Adaptive Distillation}
\label{alg:overview}
\begin{algorithmic}[1]
\REQUIRE Pretrained feature extractor $\mathnormal{f_{\theta}}$, \wnalg parameterized by $\phi_w$, \nalg parameterized by $\phi_n$, distillation coefficient function $g$, learning rate $\alpha$\\
\REQUIRE Test episode $\mathcal{T}$ = ($S$, $Q$) \\
\STATE Randomly initialize $\phi_w$, $\phi_n$\\
\textcolor{darkgray}{/* Stage2: Training \wnalg */} \\
 \WHILE{not done}
    \STATE ${L} = L_{CE}(f_{\theta, \phi_w}, S)$
    \STATE $\phi_w = \phi_w - \alpha\nabla_{\phi_w}{\mathcal{L}}(S; \theta, \phi_w)$
\ENDWHILE

\textcolor{darkgray}{/* Make the teacher features*/} \\
\textcolor{darkgray}{/* and compute the adaptive coefficient */} \\
\STATE $Y_t = f_{\theta, \phi_w}(X)$ \\
\STATE $\lambda_{\mathcal{T}} = g(f_\theta, S)$\\

\textcolor{darkgray}{/* Stage3: Domain Adative Distillation */} \\
 \WHILE{not done}
  
    \STATE ${L} = (1-\lambda_{\mathcal{T}})\cdot L_{CE}(f_{\theta, \phi_n, \phi_w}, S) + \lambda_{\mathcal{T}}\cdot L_{distill}(f_{\theta, \phi_n, \phi_w}, X, Y_t)$
   
    \STATE $\phi_n = \phi_n - \alpha\nabla_{\phi_n}{\mathcal{L}}(S; \theta, \phi_n, \phi_w)$
    \STATE $\phi_w = \phi_w - \alpha\nabla_{\phi_w}{\mathcal{L}}(S; \theta, \phi_n, \phi_w)$
\ENDWHILE
\end{algorithmic}
\end{algorithm}

\section{More Results}
\label{appendix:further_results}

In this section, we report the results that are summarized in Section \ref{sec:experiments} with additional experiments.

\subsection{Multi-Domain Learning (MDL)}

\begin{table*}[t]
\addtolength{\tabcolsep}{-2.0pt}
\centering
 \small
\begin{tabular}{ccccccccc|cc}
\toprule
        ~ & {\begin{tabular}[c]{@{}c@{}}Simple\\[-0.5ex]
CNAPS \end{tabular}} &  {\begin{tabular}[c]{@{}c@{}}Transductive\\[-0.5ex]
CNAPS  \end{tabular}}  & SUR  & URT  & FLUTE & tri-M  & URL & TSA & {\begin{tabular}[c]{@{}c@{}}\textbf{\texttt{ProLAD}}\\[-0.5ex]
\textbf{\texttt{-loss}}  \end{tabular}} & {\begin{tabular}[c]{@{}c@{}}\textbf{\texttt{ProLAD}}\\[-0.5ex]
\textbf{\texttt{-sim}}  \end{tabular}} \\
        \midrule
        ImageNet & 56.5 ± 1.1 & 57.9 ± 1.1 & 54.5 ± 1.1 & 55.0 ± 1.1 & 51.8 ± 1.1 & 58.6 ± 1.0 & 57.5 ± 1.1 & 57.4 ± 1.1 & {57.8 ± 1.1} & \textbf{59.3 ± 1.1} \\ 
        Omniglot & 91.9 ± 0.6 & 94.3 ± 0.4 & 93.0 ± 0.5 & 93.3 ± 0.5 & 93.2 ± 0.5 & 92.0 ± 0.6 & 94.5 ± 0.4  & 95.0 ± 0.4 & {95.2 ± 0.4} & \textbf{95.4 ± 0.4} \\ 
        Aircraft & 83.8 ± 0.6 & 84.7 ± 0.5 & 84.3 ± 0.5 & 84.5 ± 0.6 & 87.2 ± 0.5 & 82.8 ± 0.7 & 88.6 ± 0.5& 89.3 ± 0.4 & {89.4 ± 0.5} & \textbf{89.7 ± 0.5} \\ 
        Birds & 76.1 ± 0.9 & 78.8 ± 0.7 & 70.4 ± 1.1 & 75.8 ± 0.8 & 79.2 ± 0.8 & 75.3 ± 0.8 & 80.5 ± 0.7 & {81.4 ± 0.7} & \textbf{82.0 ± 0.8} & {81.7 ± 0.8} \\ 
        Textures & 70.0 ± 0.8 & 66.2 ± 0.8 & 70.5 ± 0.7 & 70.6 ± 0.7 & 68.8 ± 0.8 & 71.2 ± 0.8 & 76.2 ± 0.7 & 76.7 ± 0.7 & {77.8 ± 0.7} & \textbf{78.6 ± 0.7} \\ 
        Quick Draw & 78.3 ± 0.7 & 77.9 ± 0.6 & 81.6 ± 0.6 & 82.1 ± 0.6 & 79.5 ± 0.7 & 77.3 ± 0.7 & 81.9 ± 0.6 & 82.0 ± 0.6 & {82.4 ± 0.6} & \textbf{82.6 ± 0.6} \\ 
        Fungi & 49.1 ± 1.2 & 48.9 ± 1.2 & 65.0 ± 1.0 & 63.7 ± 1.0 & 58.1 ± 1.1 & 48.5 ± 1.0 & \textbf{68.8 ± 0.9}  & {67.4 ± 1.0} & {66.9 ± 1.1} & {66.4 ± 1.1} \\ 
        VGG Flower & 91.3 ± 0.6 & 92.3 ± 0.4 & 82.2 ± 0.8 & 88.3 ± 0.6 & 91.6 ± 0.6 & 90.5 ± 0.5 & 92.1 ± 0.5  & 92.2 ± 0.5 & {93.1 ± 0.4} & \textbf{93.4 ± 0.4} \\ 
        \midrule
        Traffic Sign & 59.2 ± 1.0 & 59.7 ± 1.1 & 49.8 ± 1.1 & 50.1 ± 1.1 & 58.4 ± 1.1 & 78.0 ± 0.6 & 63.3 ± 1.2  & 83.5 ± 0.9 & {87.2 ± 1.0} & \textbf{88.5 ± 0.9} \\ 
        MSCOCO & 42.4 ± 1.1 & 42.5 ± 1.1 & 49.4 ± 1.1 & 48.9 ± 1.1 & 50.0 ± 1.0 & 52.8 ± 1.1 & 54.0 ± 1.0  & \textbf{55.8 ± 1.1} & 55.2 ± 1.1 & 55.6 ± 1.1 \\ 
        MNIST & 94.3 ± 0.4 & 94.7 ± 0.3 & 94.9 ± 0.4 & 90.5 ± 0.4 & 95.6 ± 0.4 & 96.2 ± 0.3 & 94.5 ± 0.5 & 96.7 ± 0.4 & \textbf{97.6 ± 0.3} & {97.0 ± 0.3} \\
        CIFAR-10 & 72.0 ± 0.8 & 73.6 ± 0.7 & 64.2 ± 0.9 & 65.1 ± 0.8 & 78.6 ± 0.7 & 75.4 ± 0.8 & 71.9 ± 0.7  & \textbf{80.6 ± 0.8} & {79.5 ± 0.9} & {80.4 ± 0.9}\\ 
        CIFAR-100 & 60.9 ± 1.1 & 61.8 ± 1.0 & 57.1 ± 1.1 & 57.2 ± 1.0 & 67.1 ± 1.0 & 62.0 ± 1.0 & 62.6 ± 1.0  & 69.6 ± 1.0 & \textbf{71.7 ± 1.0} & {71.4 ± 1.0} \\ 
        \midrule
        Avg. seen & 74.6 & 75.1 & 75.2 & 76.7 & 76.2 & 74.5 & 80.0 & 80.2 & {80.6} & \textbf{80.9} \\ 
        Avg. unseen & 65.8 & 66.5 & 63.1 & 62.2 & 69.9 & 72.9 & 69.3 & 77.2 & {78.2} & \textbf{78.6} \\ 
        Avg. all & 71.2 & 71.8 & 70.5 & 71.1 & 73.8 & 73.9 & 75.9 & 79.0 & {79.7} & \textbf{80.0} \\
        \bottomrule
    \end{tabular}
\caption{Comparison to state-of-the-art methods on Meta-Dataset with the multi-domain setting where the feature extractor is trained on 8 datasets and then test on all datasets. Mean accuracy, 95 confidence interval are reported. }
\label{table:mdl_full}
\end{table*}

Table \ref{table:mdl_full} provides detailed results for each dataset in the MDL setting. As previously discussed, both of our methods outperform most domains, although not as much as in the SDL setting. Our method exhibits significant performance improvement in domains such as Traffic Sign (+5.0\%), which is highly dissimilar to the source set. Additionally, our method outperforms the state-of-the-art in some seen domains, such as ImageNet (+1.9\%), and in unseen domains with high domain similarity, such as CIFAR-100 (+2.1\%), indicating that the adaptive coefficient is also helpful for similar domains. Overall, \lalg and \simalg surpass other methods on 10 out of 13 domains in the less challenging MDL setting, demonstrating their effectiveness.

\subsection{Efficacy of Adaptive Coefficient on Every Domain}

\begin{figure*}[htp]
    \centering
    \includegraphics[width=1.0\textwidth]{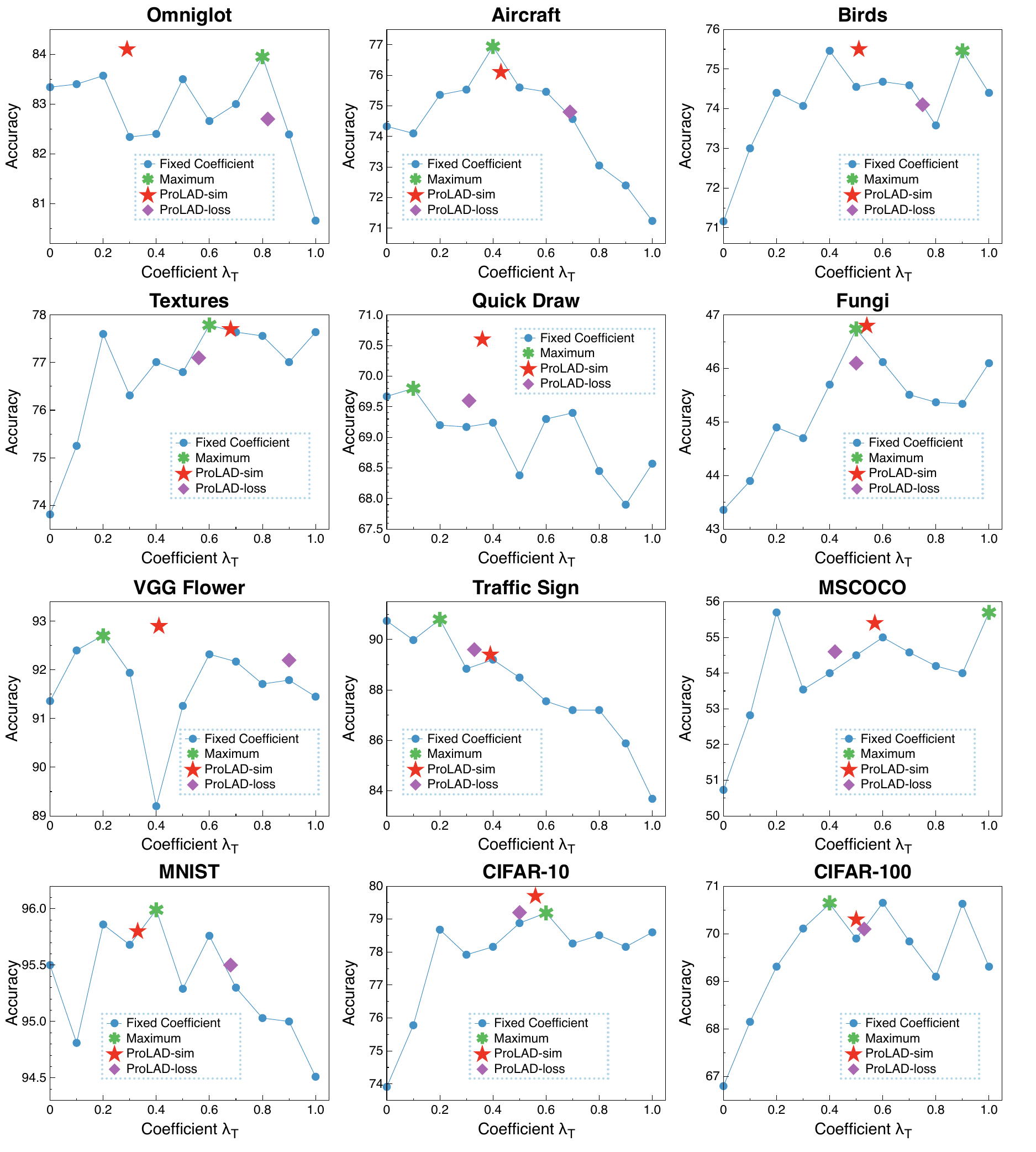}
    \caption{Performance by coefficient $\lambda_{\mathcal{T}}$ on each domain. The green mark denotes the maximum value point with a fixed coefficient. The red and purple marks represent the performance using Eq.~\eqref{eq:coeff_sim} and Eq.~\eqref{eq:coeff_loss}, respectively, where the value of x is the average coefficient across all samples. Our estimated coefficient value generally aligns with either the best or the second-best coefficient of a fixed value.}
    \label{fig:6}
\end{figure*}

Figure \ref{fig:6} displays the performance on a fixed coefficient $\lambda_{\mathcal{T}}$, in addition to the average coefficients of \simalg and \lalg, serving as an extension to Figure \ref{fig:5}. We omit the results from the source (ImageNet) to evaluate the estimated coefficient for cross-domains. For many domains, the peak performance in the fixed coefficient aligns closely with the coefficient estimated by \simalg. While certain domains like Omniglot and Birds do not match the maximum point, these coefficients align with the second-highest point, showcasing the effectiveness of using Eq.~\eqref{eq:coeff_sim} for proper coefficient estimation. Additionally, while \lalg generally provides good estimates, it struggles with domains such as MNIST and VGG Flower, which exhibit high domain discrepancies but are relatively easy to classify. As a result, the coefficient is estimated too high for these domains. Overall, our estimated coefficients align well and often surpass the peak performance of a fixed coefficient, highlighting the advantages of adaptiveness.

\subsection{Ablation Study on \nalg}

\begin{table*}[t]
\centering
\scriptsize
\setlength{\tabcolsep}{1.75pt}
\renewcommand{\arraystretch}{1.8}
\begin{tabular}{lcccccccccccccc}
    \toprule
       {Architecture} & {\begin{tabular}[c]{@{}c@{}}Image\\[-2.5ex]
-net \end{tabular}} & {\begin{tabular}[c]{@{}c@{}}Omni\\[-2.5ex]
-glot \end{tabular}}& {\begin{tabular}[c]{@{}c@{}}Airc\\[-2.5ex]
-craft \end{tabular}}& {Birds}& {\begin{tabular}[c]{@{}c@{}}Tex\\[-2.5ex]
-tures \end{tabular}}& {\begin{tabular}[c]{@{}c@{}}Quick\\[-2.5ex]
Draw \end{tabular}}& {Fungi}& {\begin{tabular}[c]{@{}c@{}}VGG\\[-2.5ex]
Flower \end{tabular}}& {\begin{tabular}[c]{@{}c@{}}Traffic\\[-2.5ex]
Sign \end{tabular}} & {\begin{tabular}[c]{@{}c@{}}MS\\[-2.5ex]
-COCO \end{tabular}} & {MNIST}& {\begin{tabular}[c]{@{}c@{}}CIFAR\\[-2.5ex]
-10 \end{tabular}}& {\begin{tabular}[c]{@{}c@{}}CIFAR\\[-2.5ex]
-100 \end{tabular}} & {Avg}\\ 

        \midrule
        FiLM & 57.2 ± 1.1 & 76.8 ± 1.3 & 72.3 ± 1.2 & 75.0 ± 0.9 & \textbf{77.9 ± 0.7} & 68.0 ± 1.0 & 45.7 ± 1.2 & 92.1 ± 0.6 & 84.1 ± 0.9 & 54.9 ± 1.1 & 94.2 ± 0.6 & \textbf{80.0 ± 0.7} & \textbf{71.4 ± 0.9} & 73.0    \\
        Conv & 57.2 ± 1.1 & 80.5 ± 1.2 & 74.8 ± 1.1 & 74.4 ± 1.0 & 76.6 ± 0.8 & 68.0 ± 0.9 & 46.1 ± 1.2 & 92.4 ± 0.6 & 86.8 ± 1.0 & 53.6 ± 1.1 & 95.2 ± 0.5 & 78.6 ± 0.9 & 69.6 ± 1.1 & 73.4    \\
         GroupConv & {55.9 ± 1.1} & 78.7 ± 1.2 & 72.8 ± 1.2 & {75.0 ± 1.0} & {77.6 ± 0.8} & 68.5 ± 1.0 & 45.3 ± 1.2 & 91.6 ± 0.6 & 84.1 ± 0.9 & 55.5 ± 1.1 & 94.5 ± 0.5 & 79.7 ± 0.8 & 70.5 ± 0.9 & 72.6  \\ 
         SN + Conv & 56.0 ± 1.1 & 82.2 ± 1.2 & 75.4 ± 1.2 & \textbf{75.7 ± 1.0} & 77.1 ± 0.8 & 69.8 ± 0.9 & {45.7 ± 1.1} & {92.4 ± 0.6} & 89.0 ± 0.9 & 54.6 ± 1.0 & 95.7 ± 0.6 & {79.5 ± 0.8} & 70.0 ± 1.0 & 74.1   \\ 
         SN + FiLM (BN) & 57.1 ± 1.1 & 78.7 ± 1.2 & 72.8 ± 1.2 & 75.0 ± 1.0 & 77.6 ± 0.8 & 68.5 ± 1.0 & 45.3 ± 1.2 & 91.6 ± 0.6 & 84.1 ± 0.9 & \textbf{55.5 ± 1.1} & 91.6 ± 0.6 & 79.7 ± 0.8 & 70.5 ± 0.9 & 73.1   \\ 
         SN + GroupConv (Ours) &  \textbf{57.2 ± 1.1} & \textbf{84.1 ± 1.2}& \textbf{76.1 ± 1.2}& 75.5 ± 1.0 & 77.7 ± 0.8 & \textbf{70.6 ± 1.0} & \textbf{46.8 ± 1.2}& \textbf{92.9 ± 0.6}& \textbf{89.4 ± 0.9}& {55.4 ± 1.1} & \textbf{95.8 ± 0.5}& {79.7 ± 0.8}& {70.3 ± 1.0 }&\textbf{74.7}  \\ 
        \bottomrule
    \end{tabular}
\caption{Ablation study on the components of \nalg. Mean accuracy, 95 confidence interval are reported.} 
\label{table:ablation_design_full}
\end{table*}

Detailed results of the ablation study on \nalg, evaluated using \simalg, are presented in Table \ref{table:ablation_design_full}. Overall, our design choice surpasses alternative options in 8 out of 13 domains, showcasing its effectiveness when combined with adaptive distillation. Additionally, employing a group convolutional network yields superior performance compared to using a 1x1 convolutional network, especially when combined with a normalization layer. This suggests that a larger receptive field offers benefits for feature modulation, particularly when both network types have a similar number of parameters. Experiments involving the FiLM layer reveal that adapters represented solely by a vector are not sufficient; more robust adapters parameterized by a matrix are required. As previously mentioned, integrating standard normalization is especially beneficial when domain similarity is low, underscoring our hypothesis about the importance of including the batch statistics from the target domain.

\subsection{Ablation Study on Scaling Factor $\beta$}

\begin{figure}[ht]
    \centering
    \includegraphics[width=1.0\columnwidth]{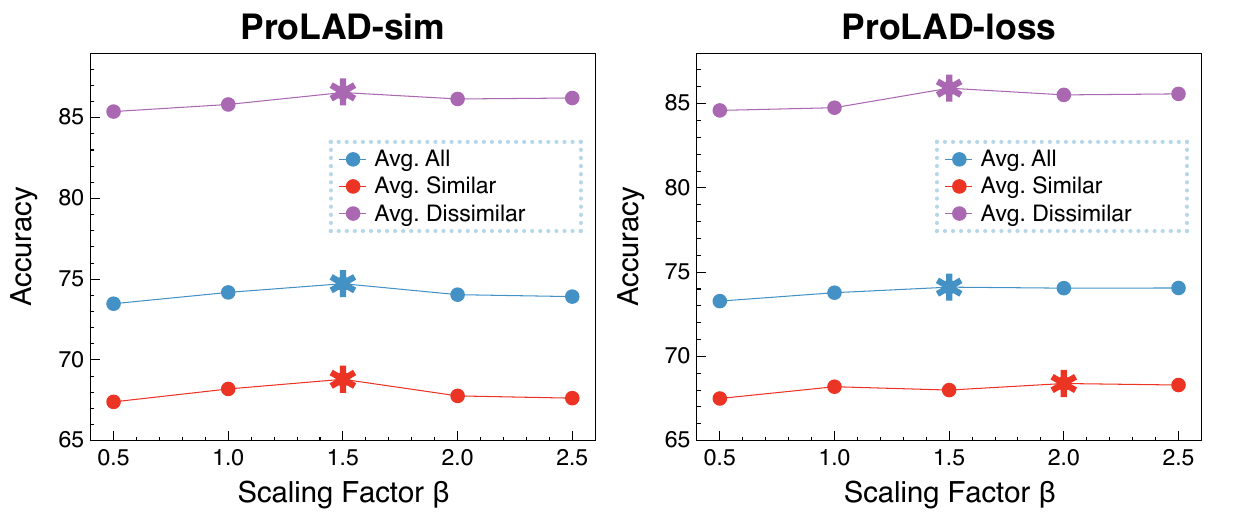}
    \caption{Performance of \alg based on the scaling factor $\beta$. The maximum performance for each method is denoted with an asterisk mark.}
    \label{fig:7}
\end{figure}

In this section, we examine the robustness of our method with respect to the hyperparameter of the scaling factor, $\beta$, as illustrated in Figure \ref{fig:7}. We have conducted a hyperparameter search in the range of 0.5 to 2.5, since values greater than 2.5 or less than 0.5 result in a fixed coefficient of either 0.0 or 1.0, respectively. Our findings indicate that our method is not overly sensitive to hyperparameters, especially in the 1.0 to 2.0 range. This suggests that our method is robust, and the adaptability and tendency to adjust based on domain similarity is more crucial than the absolute value, as long as these values are neither too small nor too large. As the overall performance peaks at 1.5 for both \simalg and \lalg, we have chosen 1.5 as the scaling factor for our main experiments.

\section{Additional Related Work}
\subsection{Batch Normalization for Domain Adaptation and Domain Generalization}

Due to the distribution discrepancy between the source and target domains, directly applying batch normalization trained from the source dataset can impair performance on the target in domain adaptation and domain generalization \cite{li2016revisiting, bilen2017universal}. \citet{chang2019domain} proposed a method involving a domain-specific batch normalization layer using two branches, while \citet{wang2019transferable} introduced transferable normalization. This challenge has also been addressed in the context of domain generalization. For instance, \citet{seo2020learning} proposed learning to optimize domain-specific normalization for domain generalization. More recently, \citet{mirza2022norm} highlighted the challenge of adapting batch normalization with limited data and proposed online adaptation of batch normalization. Meanwhile, addressing shifts in batch statistics has also been applied to few-shot scenarios, combined with meta-learning to learn adaptive statistics \cite{du2020metanorm}.

\subsection{More Cross-Domain Few-Shot Learning Methods}

In this section, we explore approaches to address CD-FSL that are not presented in Section \ref{sec:related_work}. Some methods employ distinct techniques to tackle the problem. For instance, CTX \cite{doersch2020crosstransformers} proposes learning spatial correspondences from the base classes and evaluating them on the novel classes, while BOHB \cite{saikia2020optimized} leverages hyperparameter optimization to enhance generalization capability. \citet{qin2023bi} adopts a two-stage meta-learning method for few-shot domain generalization: first learning intra-domain knowledge, followed by inter-domain knowledge.

Other approaches presume the availability of unlabeled data from the test domain, which steers them towards unsupervised learning techniques. These methods include distillation \cite{islam2021dynamic} or self-supervised learning \cite{phoo2020self, chen2020simple} for pretraining. However, in this work, we operate under the assumption that unlabeled data from the test domain is unavailable during pretraining, considering this a more practical scenario.

\end{document}